\documentclass[journal]{IEEEtran}

\usepackage[francais,english]{babel}
\usepackage[applemac]{inputenc}
\usepackage{amsmath,graphicx,url,multirow}
\usepackage[table]{xcolor}
\usepackage{color}
\usepackage{url}

\def\etal{{\textit{et al. }}}

\begin{document}

\title{A Three-Feature Model to Predict Colour Change Blindness}

\author{Steven Le Moan, Marius Pedersen
\thanks{Steven Le Moan is with the School of Food and Advanced Technology at Massey University, Palmerston North, New Zealand (e-mail: s.lemoan\at massey.ac.nz)}}

\maketitle

\begin{abstract}
Change blindness is a striking shortcoming of our visual system which is exploited in the popular 'Spot the difference' game. It makes us unable to notice large visual changes happening right before our eyes and illustrates the fact that we see much less than we think we do. We introduce a fully automated model to predict colour change blindness in cartoon images based on two low-level image features and observer experience. Using linear regression with only three parameters, the predictions of the proposed model correlate significantly with measured detection times. We also demonstrate the efficacy of the model to classify stimuli in terms of difficulty.
\end{abstract}



\setcounter{section}{-1} 
\section{Introduction}
\label{sec:intro}

\begin{figure}[!ht]
	\begin{center}
            \begin{tabular}{cc}
             \includegraphics[width=0.46\linewidth]{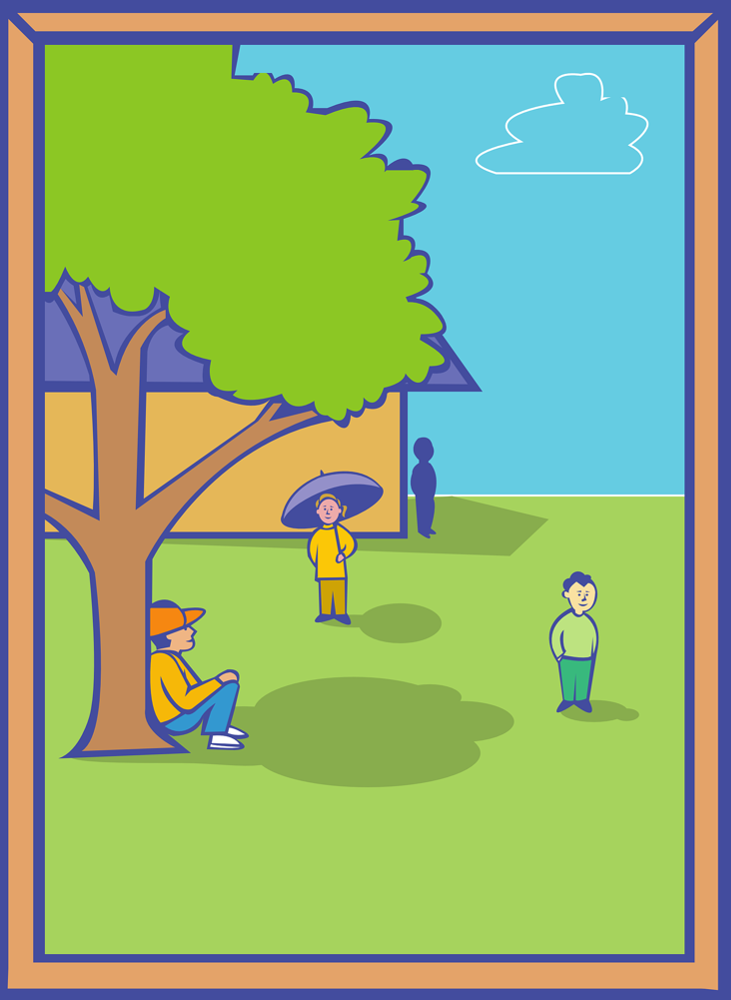} & 
             \includegraphics[width=0.46\linewidth]{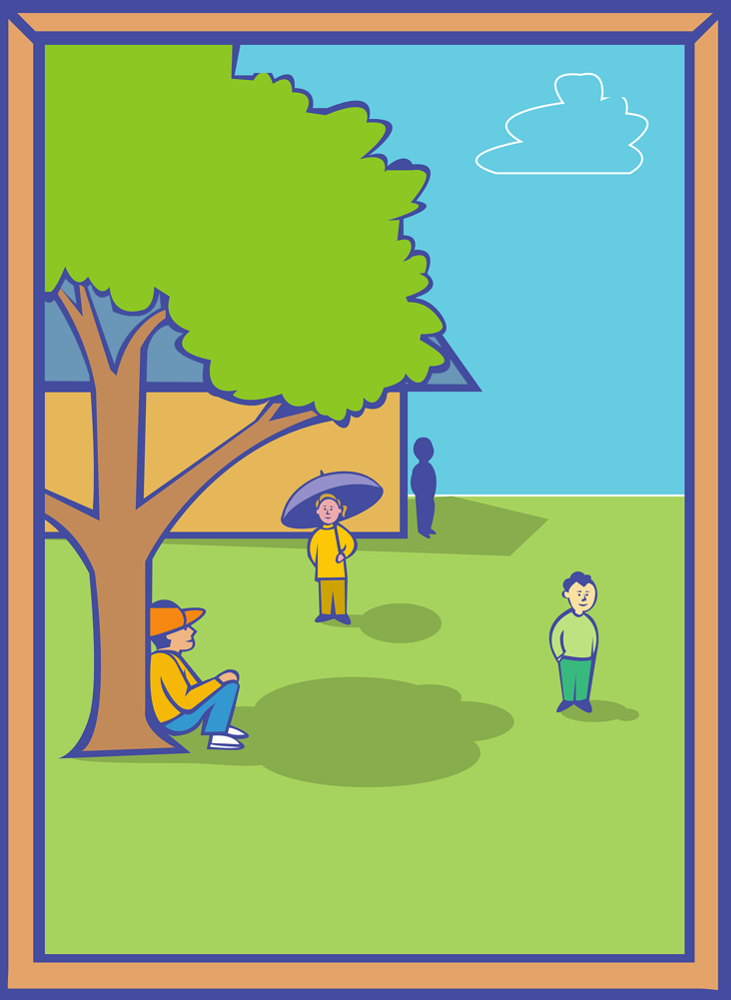} \\
            \end{tabular}
        \caption{Example of change blindness-inducing image pair.}
    \label{fig:example1}
	\end{center}
\end{figure}

Despite our impression of a richly detailed visual world, our perceptual experience is surprisingly limited. For instance, consider the 'Spot the difference' game in Fig. \ref{fig:example1}. We found that most people need at least 13 seconds to notice the change under normal viewing conditions (see footnote for the solution\footnote{The roof has a different colour.}). This is an example of \emph{change blindness}, a striking shortcoming of our visual system caused by limitations of attention and memory \cite{simons1997change,jensen2011change,cohen2016bandwidth}. Despite what its name may suggest, change blindness is not a disability, nor is it due to damage to the visual system: everyone is subject to it. It also tends to be significantly underestimated \cite{loussouarn2011exploring}, i.e. we see far less than we think we do.

On the one hand, change blindness is known to be responsible for accidents in car traffic \cite{galpin2009change,hallett2011cell}, process plants \cite{liu2017identifying} and even submarines \cite{NYT1}, and also for unreliable eyewitness testimony \cite{davies2007change}. On the other hand, it has recently been shown that this perceptual failure can be harnessed to reduce computational load in computer graphics \cite{cater2003varying}, to design virtual realities \cite{suma2011leveraging}, for image compression \cite{le2018towards} and visual quality assessment \cite{le2018towards}. Change blindness prediction is also useful in visualisation and user interfaces \cite{wilson2018autopager,brock2018change}. However, the cognitive mechanisms responsible for change blindness are not well understood, which has limited the ability to develop a robust model, particularly for computer graphics and image processing applications.

In this paper, we propose a fully automatic model to predict detection times (DTs) in image pairs where a single change in colour is introduced during a flicker. The model is based on low-level features that represent the complexity of the scene as well as the magnitude of the change. We also demonstrate that user experience, estimated as the number of stimuli previously viewed, correlates significantly with DTs and we include it in our model. Data was collected via a user study with 66 participants in two different locations: New Zealand and Norway. We used cartoon images as stimuli as they allow to obtain more robust scene and object attributes that are not reliant on noisy low-level feature extractors \cite{zitnick2016adopting}. A total of 3418 valid DTs were recorded.

After reviewing related work, we describe our experimental design and analyse the results obtained in terms of intra- and inter-observer variability. We then present our model and evaluate its ability to predict DTs as well as to classify image pairs as either easy or hard.

\section{Related work}

Early accounts of people's inability to notice unexpected visual changes in their environment date back to the 19th century\cite{james1890principles}. Yet, it was not before the early 1990s that change blindness research gained real traction, following decades of related work on eye movements and visual memory\cite{simons1997change}.

Change blindness is a form of high-level visual masking, whereby the change is masked by a disruption and by visual clutter. It is different from low-level masking\cite{alam2014local,lavoue2018psychophysical} which prevents the perception of a target (e.g. a compression artefact) even though the observer knows where it is, mostly due to limitations of early vision. The disruption that induces change blindness can be abrupt (flicker\cite{simons1997change}, saccade\cite{henderson2003global}, 'mudsplashes'\cite{o1999change}, motion change\cite{yao2019if}) or gradual\cite{simons2000change}. Either way, it is key to preventing exogenous orienting to the change due to bottom-up salience. Instead of using a disruption, one can purposely direct the observer's attention away from the target to induce another kind of perceptual failure known as inattentional blindness. In a famous experiment\cite{simons1999gorillas}, Simons \etal{} showed that a person dressed as a gorilla walking among people playing basketball can go completely unnoticed if the observer is focused on counting the number of passes between players. Change blindness and inattentional blindness are both failures of visual awareness. However, they each have a unique background and distinct theoretical implications\cite{jensen2011change}.

The specific causes of change blindness are still debated, but they are known to be linked with attention and memory \cite{tseng2010posterior,cohen2016bandwidth}. Visual input data is about 1GB/s \cite{kelly1962information} and it is compressed in the retina to about 8.75MB/s of raw information \cite{koch2006much}, which is substantially more than what the brain can handle \cite{sziklai1956some}. Therefore, sensory signals undergo various stages of transformation and selection so they can be processed and interpreted efficiently.

In the visual cortex, spatial pooling over receptive fields of increasing size, latency and complexity\cite{freeman2011metamers} is akin to a lossy signal compression, where only the most important information is conveyed from one level to the next. Recent advances in cognitive science\cite{freeman2011metamers}, brain imaging\cite{pessoa2004neural} and machine learning\cite{yamins2014performance} have led to a better understanding of the higher visual cortex, yet there is currently no consensus as to exactly where and how information loss occurs during change blindness \cite{cohen2016bandwidth}. Some researchers support the idea that visual information is only consciously perceived once it is accessed by the attentional functions of late vision, so the cognitive mechanisms associated with awareness, attention, memory and decision-making are intrinsically linked. Others argue that signals emanating from early vision can all reach consciousness but are only partially accessible by the mechanisms associated with decision-making. Both views agree on the existence of a tight bottleneck which researchers from fields such as psychology, cognitive science and philosophy have set out to characterise for over three decades.

However, there have been only a few attempts at creating a computational model able to predict high-level visual masking in natural scenes. Existing change blindness models \cite{le2018towards,MA13,jin2013spoid,HOU12,verma2010semi} based on bottom-up salience and image segmentation are \textit{ad hoc} and rely on parameter tuning which is challenging given the paucity of available reference data1, particularly for applications related to visual quality.

Change blindness depends on individual factors such as observer experience \cite{werner2000change}, age \cite{rizzo2009change} and culture \cite{masuda2006culture}. A recent study \cite{andermane2019individual} revealed that detection performance is associated with the ability to form stable perceptual inferences and with being able to resist task-irrelevant distractors. A battery of tests were used to characterise the idiosyncrasy of change blindness in terms of cognitive and attentional capacities. However, for most practical applications of change blindness, such tests are particularly tedious. In this paper, we are interested in predicting change blindness based on stimulus-related factors and limited information about the user.

It has been shown that the masking effect is stronger in visually complex scenes, as they are difficult to encode and maintain in visual short-term memory \cite{alvarez2004capacity,tseng2010posterior,cohen2016bandwidth}. Changes that affect the gist of the scene \cite{zuiderbaan2017change}, or which significantly affect bottom-up salience \cite{verma2010semi} are detected more rapidly. Hou \etal \cite{HOU12} proposed a stimulus-driven predictive model of change blindness based on frequency domain-based salience detection. They compared the sign of DCT coefficients in the original and changed images and found that the number of unequal signs correlates with DTs. Effectively, this approach allows estimating the imbalance in salience between the two images. If the changed object is significantly more salient in one image than in the other, the change itself becomes salient and easier to detect. Verma \etal \cite{verma2010semi} developed a semi-automated approach to generate image pairs with a desired degree of change blindness, based on salience imbalance. Note also that salience imbalance has been shown to predict subjective assessments of image quality in pair comparison setups \cite{ZHAN14}. Several models have been proposed based on the same precepts, using image segmentation and bottom-up saliency to predict change blindness in a semi-automated \cite{jin2013spoid} or fully automated way \cite{MA13}. However, these methods rely on many intrinsic parameters requiring user input or high-dimensional parameter optimisation. Furthermore, they do not account for the bias due to user experience. Indeed, users tend to become better at spotting the difference after practising on a few examples and this short-term experience predicts change detection performance.

Another gap in the current state-of-the-art is the lack of available reference data to study change blindness. Sareen \etal \cite{sareen2016cb} produced a database of 109 image pairs observed by 24 participants, but the order in which stimuli were displayed to individual observers is not provided.

We carried out an extensive user study under controlled conditions, which involved 66 participants in two locations: New Zealand and Norway. This paper has two contributions:
\begin{itemize}
    \item A new publicly available benchmark for change blindness research which includes scene rank.
    \item A new predictive model of change blindness based on low-level image features and user experience (scene rank).
\end{itemize}


\section{User study}

\subsection{Participants and Instructions}

A total of 66 participants were recruited: 40 in Palmerston North, New Zealand and 26 in Gj{\o}vik, Norway. Age ranged from 18 to 65 (median: 34.5) and 71\% of them were male. All had normal or adjusted-to-normal vision and passed an Ishihara test for colour blindness. They all declared having at least a good command of the English language.

Each observer signed a consent form, as per Massey University's Research Ethics guidelines. They were then instructed as follows (in English):\\

\begin{itemize}
    \item \textit{Your task is to spot the differences in pairs of images which will be displayed on the screen. Each pair contains a single difference.}
    \item \textit{Unlike what you may expect, images will not appear side by side. They will be shown one after the other in a flickering fashion. As soon as you spot the difference, please click on it as quickly as possible. You can click only once. After clicking, the solution will be displayed whether you were correct or not. If after one minute you have not noticed any difference, the solution will be displayed anyway and it will move to the next pair.}
    \item \textit{After each sequence of five images, you will have the opportunity to take a short break. Make it as long as you need and feel free to stop the experiment at any time, especially if you feel your focus is drifting away from the task.}
\end{itemize}

The purpose of the latter instruction is to minimise subjective bias due to fatigue, low engagement and negative emotions \cite{porubanova2013change}. Participants viewed 61 stimuli on average (min: 10, max: 100). Image sequences were randomised under two constraints:

\begin{itemize}
    \item All stimuli had to be seen by an approximately equal number of observers.
    \item Some scenes had to be seen earlier than others so that the average rank (position in sequence) across all observers follows approximately a uniform distribution. This allowed studying the influence of experience in change detection performance.
\end{itemize}

\subsection{Apparatus}

We used Eizo ColorEdge displays (CG2420 in New Zealand and CG246W in Norway), both 61cm/24.1" and calibrated with an X-Rite Eye One spectrophotometer for a colour temperature of 6500K, a gamma of 2.2 and a luminous intensity of 80cd/m$^2$. Both experiments were carried out in a dark room. The distance to the screen was set to approximately 50cm (without chin rest). Observers were reminded to keep a constant distance to the screen at the beginning and, when needed, during the experiment.

\subsection{Stimuli}

\begin{figure*}[!ht]
	\begin{center}
            \begin{tabular}{ccccc}
             \includegraphics[width=0.15\linewidth]{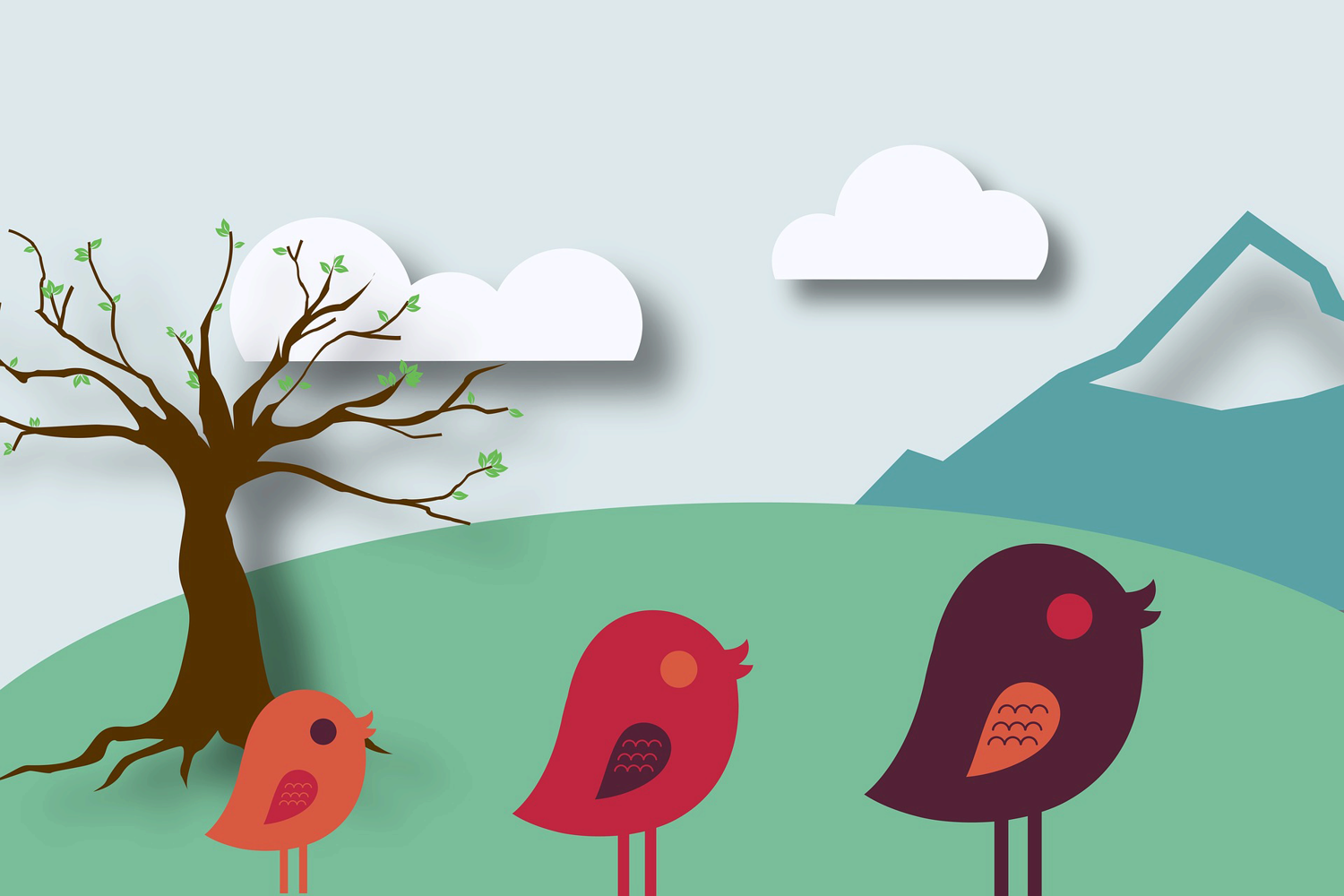} & 
             \includegraphics[width=0.15\linewidth]{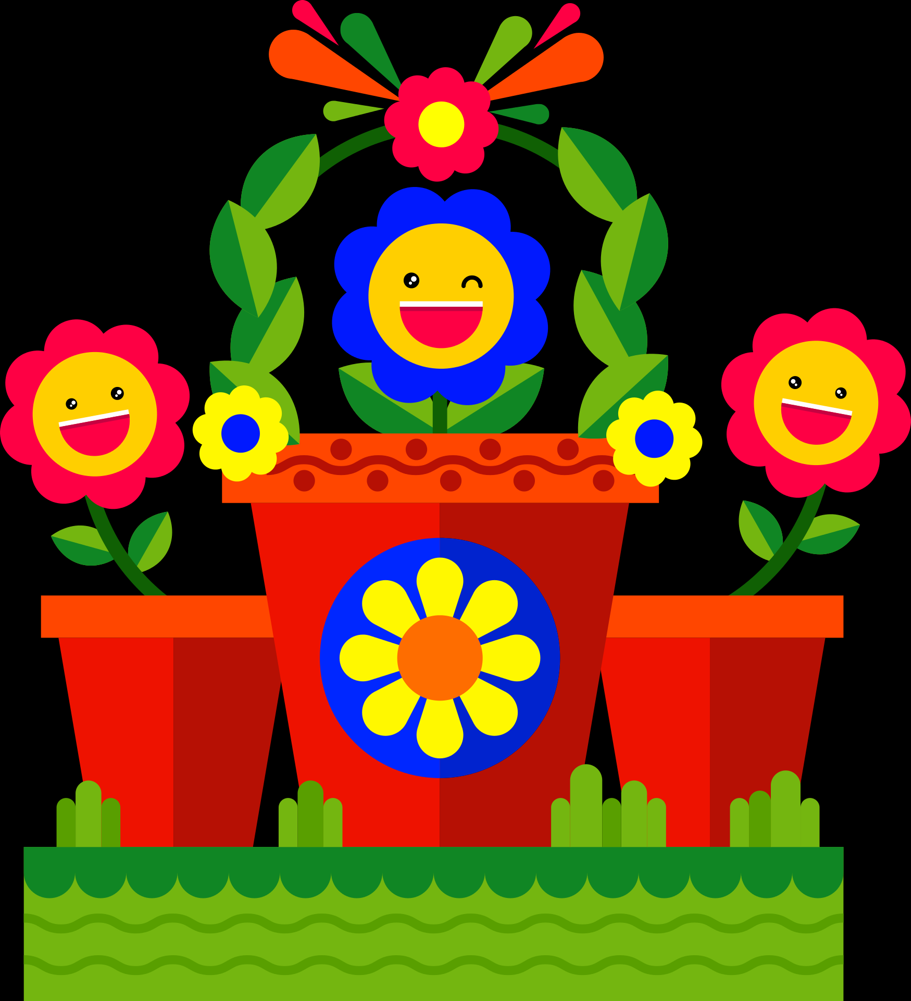} & 
             \includegraphics[width=0.15\linewidth]{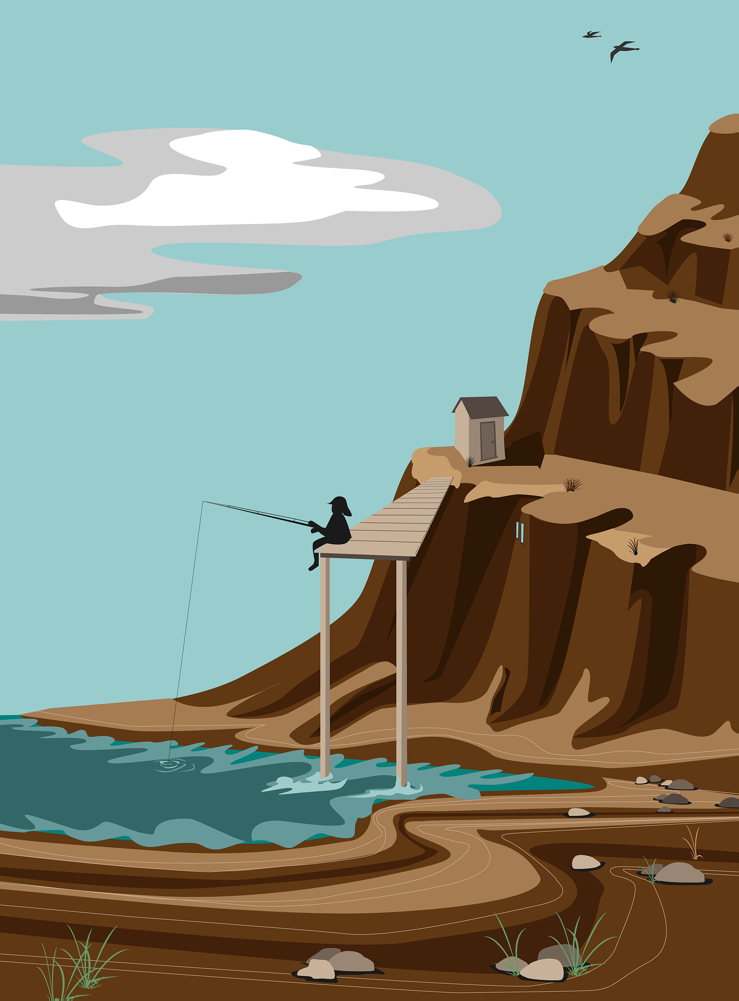} & 
             \includegraphics[width=0.15\linewidth]{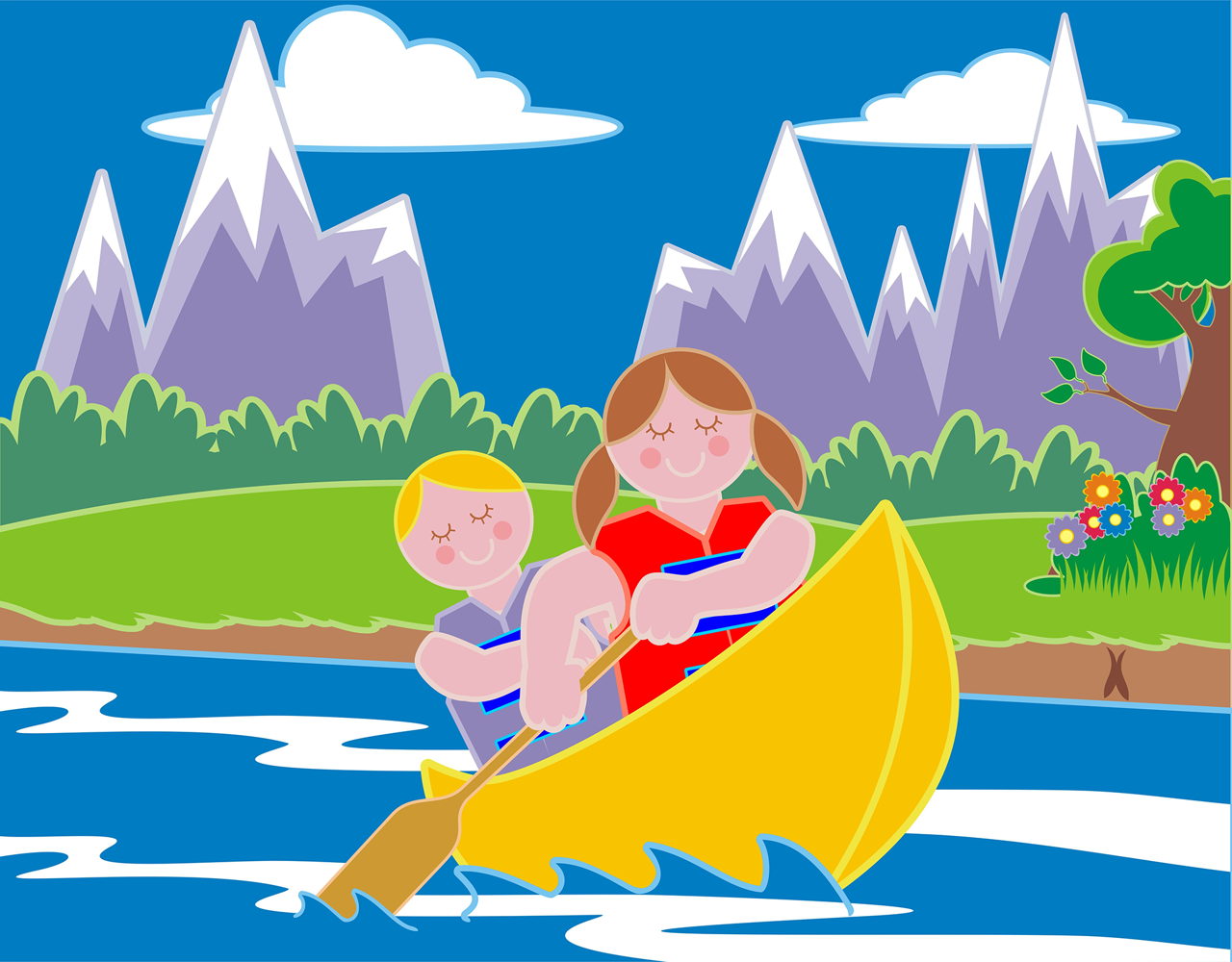} & 
             \includegraphics[width=0.15\linewidth]{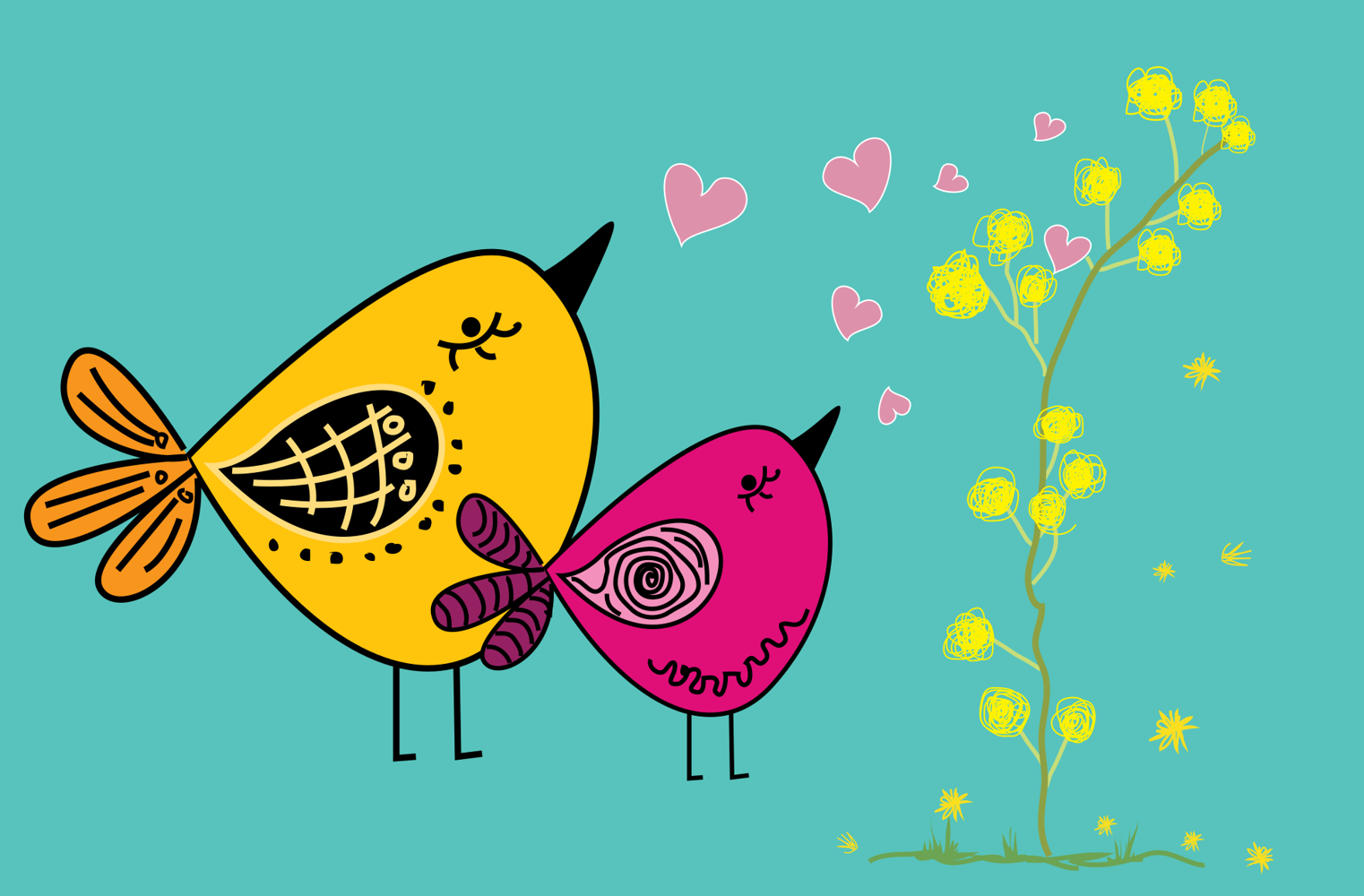} \\
             \includegraphics[width=0.15\linewidth]{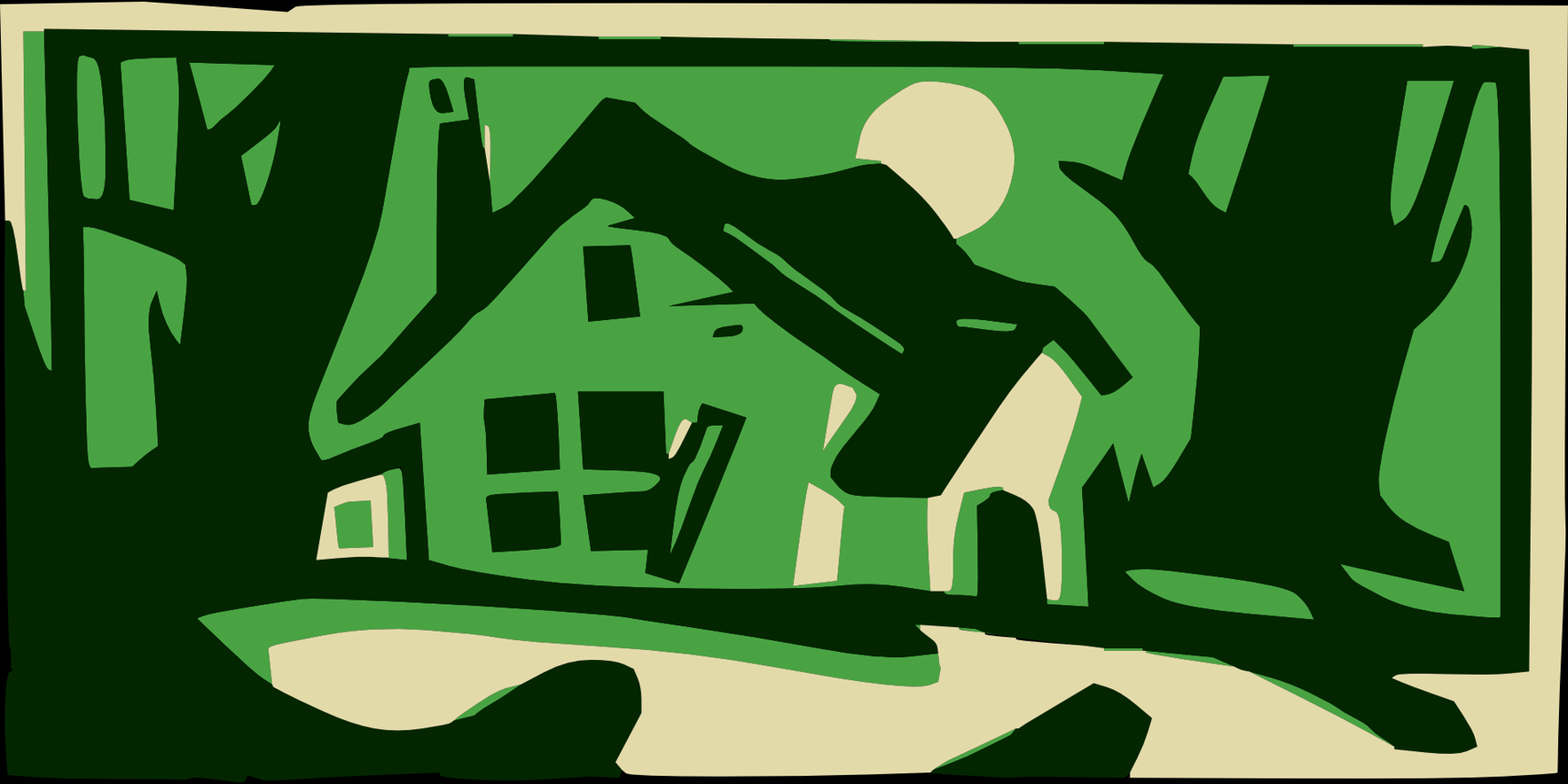} & 
             \includegraphics[width=0.15\linewidth]{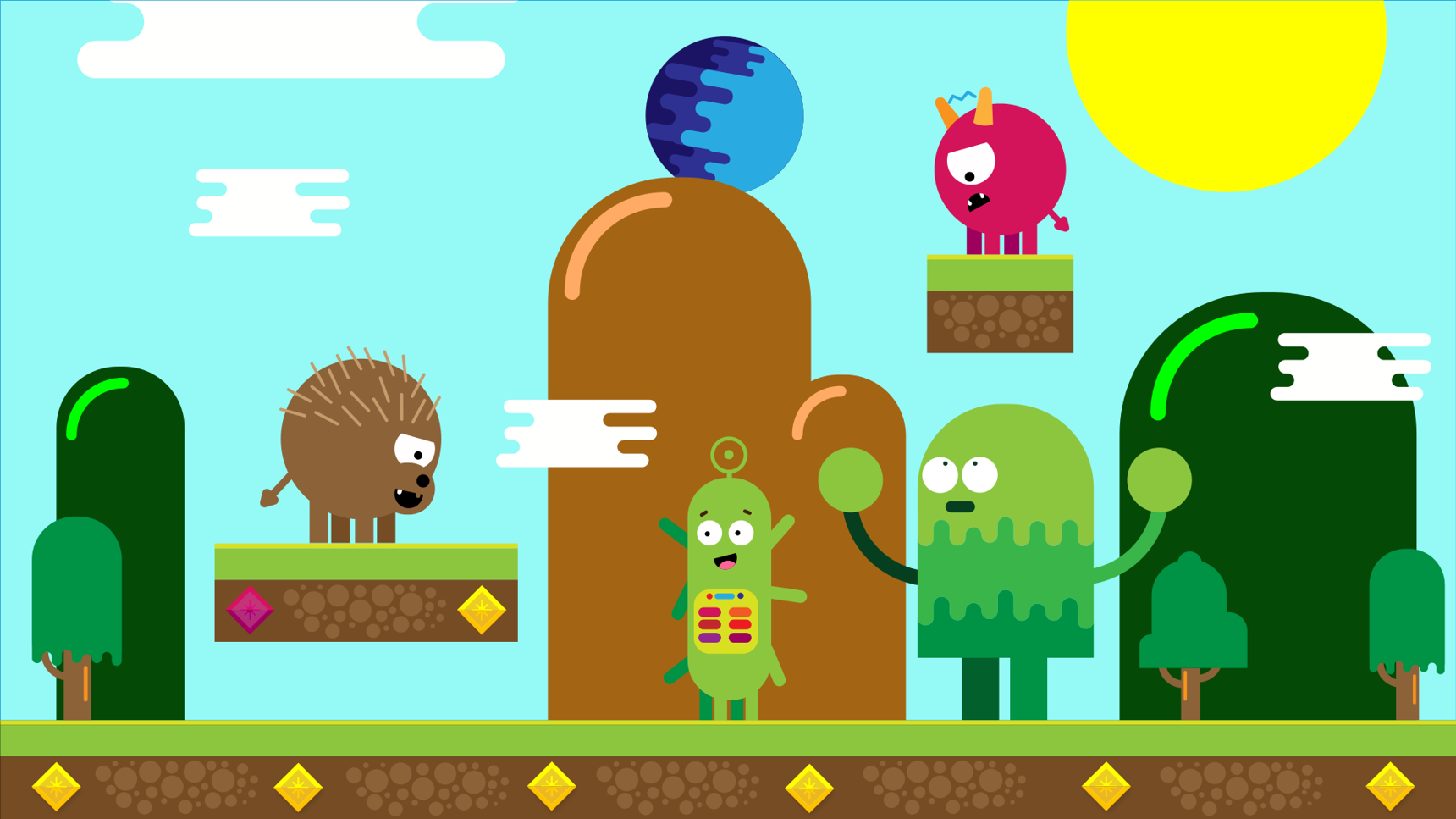} & 
             \includegraphics[width=0.15\linewidth]{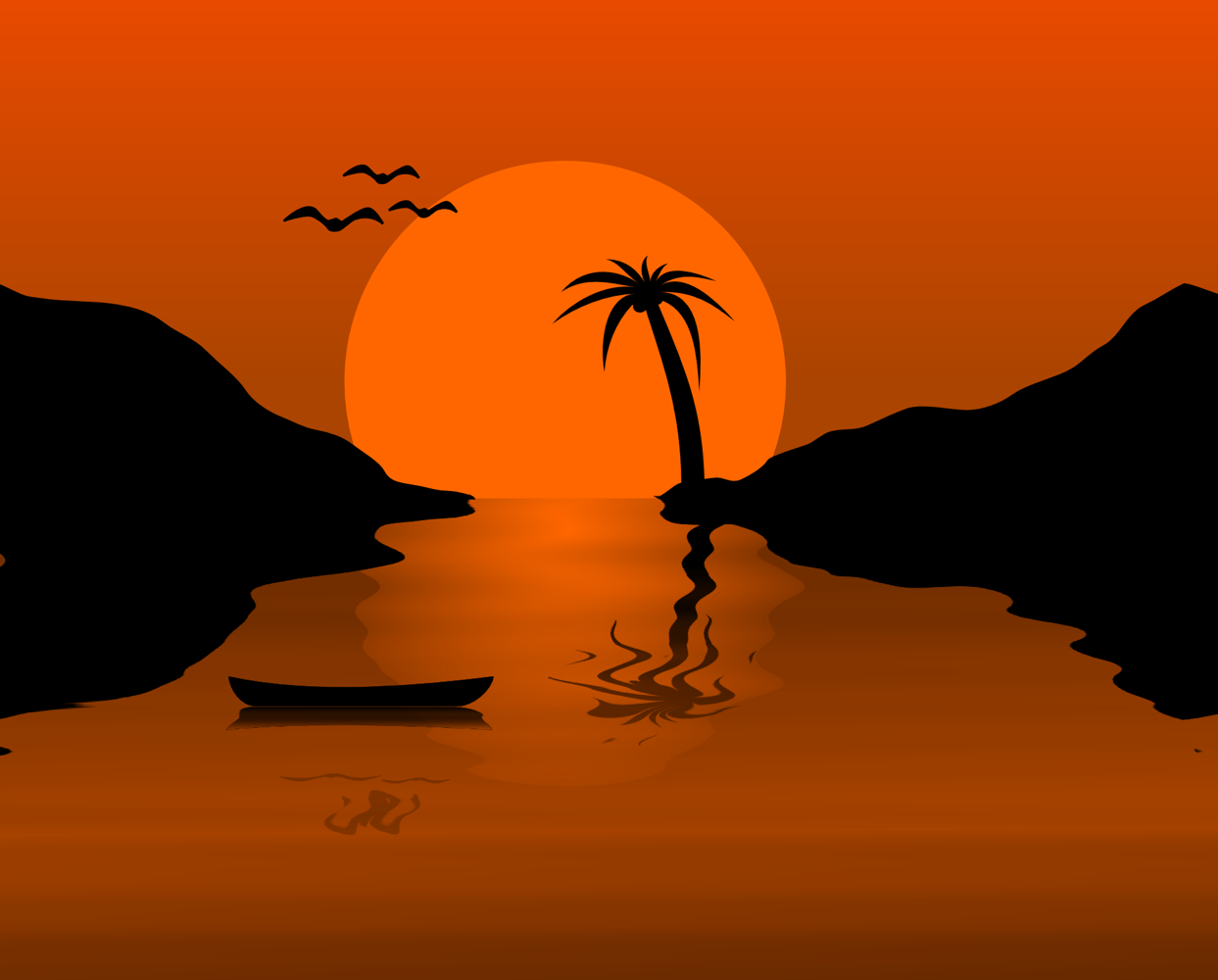} & 
             \includegraphics[width=0.15\linewidth]{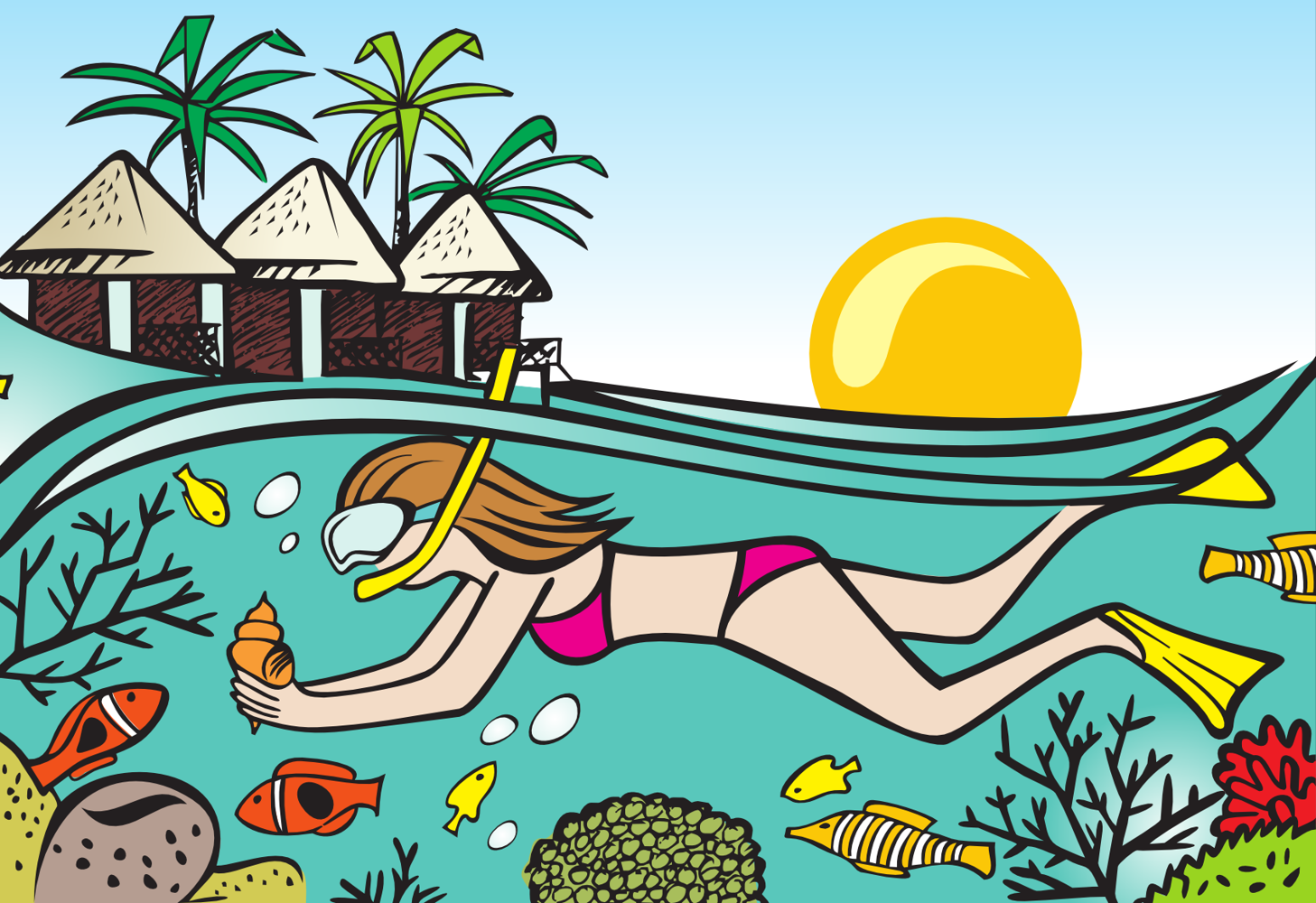} & 
             \includegraphics[width=0.15\linewidth]{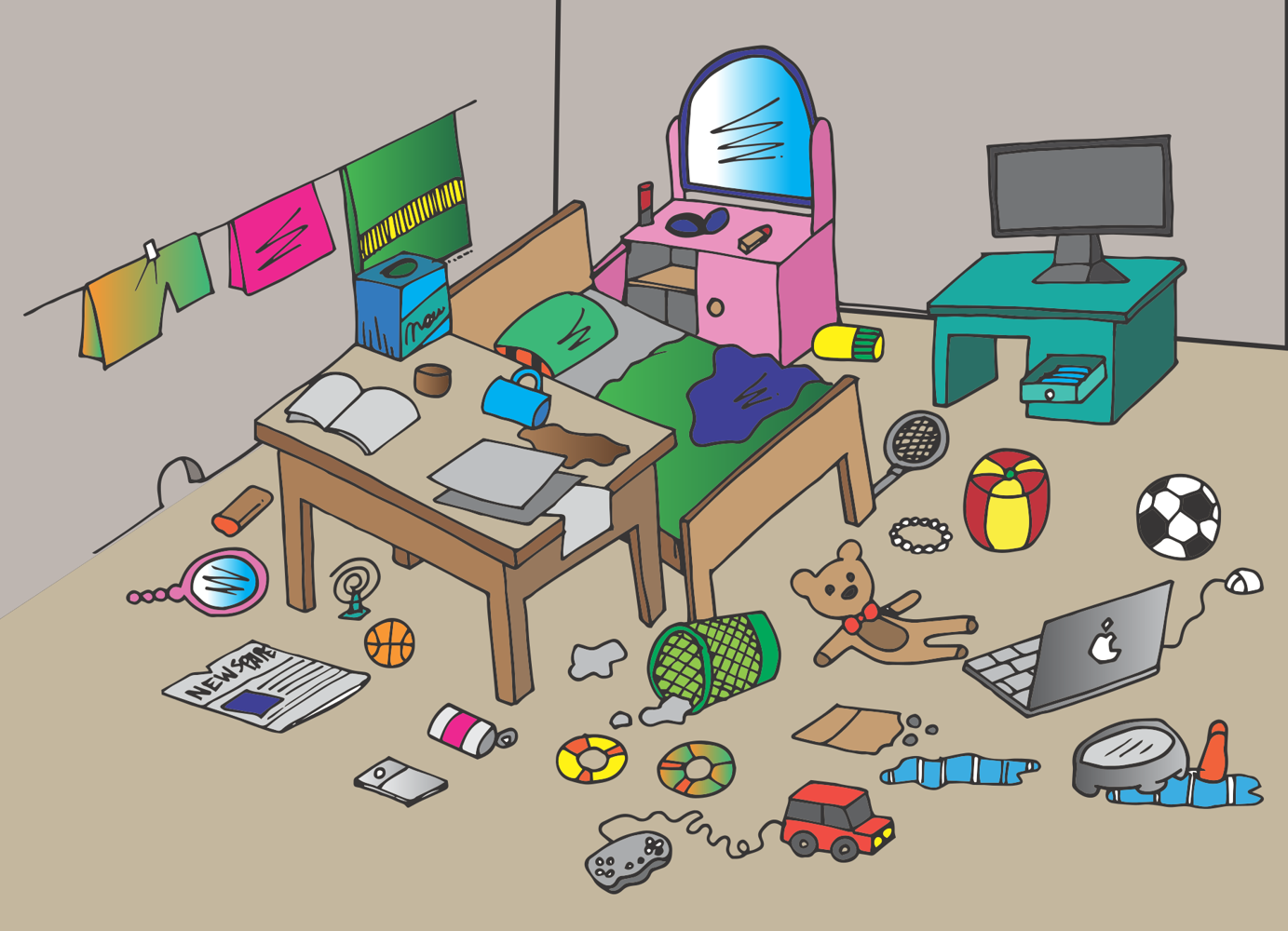} \\
            \end{tabular}
        \caption{Example scenes from our benchmark}
    \label{fig:examples3}
	\end{center}
\end{figure*}

In total, 100 copyright-free colour images were selected from various online image databases (see Figure \ref{fig:examples3}). We chose exclusively cartoon scenes, as they have the advantage to contain clear, sharp object boundaries, thus making it easier to make seamless changes to specific objects in the scenes. Furthermore, in order to reduce top-down bias due to familiarity with the scenes depicted in the stimuli, we chose them so that they contain no recognisable text, landmarks, characters or cultural symbols of any sort. We also aimed to have as uniform a variety of scene complexity (measured objectively with sub-band entropy \cite{rosenholtz2007measuring} and edge density \cite{mack2004computational}) as possible. Images also vary in terms of size, with a minimum resolution of 640x1000 and a maximum of 1720x970. The latter fits on the display with a 100-pixel grey margin on the left and right.

Alternative versions of each of these scenes were created with an image-editing software. The changes made were only in terms of colour (as opposed to moving or removing an object). They affect a single object in each scene, but that object is not necessarily compact as it can be made of several components, e.g. due to occlusion. Changes are in either hue, chroma, lightness, or any combination. The minimum magnitude of change was set to 1.2 units of Euclidean distance (median: 5.3, max: 22.3) in the perceptually uniform and hue-linear LAB2000HL colour-space \cite{LissnerPreissUrban2011}, where one unit corresponds approximately to the threshold of just noticeable difference based on the display's RGB space. The average magnitude over all pairs is about seven units. Changes affect objects of a balanced variety of size, eccentricity and importance (background/foreground). Fig. \ref{fig:examples3} shows examples of stimuli used in the study. The original and modified scenes are respectively noted $\mathbf{s}_i$ and $\mathbf{s}_i^\ast$ ($i=1,\dots,100$). Each pair $\mathbf{p}_i=\left\{\mathbf{s}_i, \mathbf{s}_i^\ast\right\}$ is associated with a decision time for observer $o$ noted $T_{o,i}$.

Change blindness was then induced by means of the flicker paradigm \cite{simons1997change}, with an 800 ms display time and a 80 ms flicker. The stimuli $\mathbf{s}_i$ and $\mathbf{s}_i^\ast$ were displayed successively for a maximum duration of 60 s or until the change was noticed. Whether the first image was $\mathbf{s}_i$ or $\mathbf{s}_i^\ast$ was decided randomly at each trial.

For each observer, we generated a unique pseudo-random sequence of image pairs. As previously mentioned, the randomness was controlled so as to compensate for the fact that not all observers stayed for the whole set. This led to the collection of 34 times per $\mathbf{p}_i$ on average (min: 21, max: 41) after observer screening (see below).

\subsection{Screening}

Observers were screened based on false positive rate (when a change was reported but far from the changed region) and non-detection rate (when no change was found in less than 60 seconds). In both cases, any observer with a rate larger than the average plus two standard deviations was considered an outlier. In total, six participants were discarded, three of whom had too high a false positive rate. From visual inspection of the recorded mouse clicks of these participants (recall that observers were asked to click where they saw a change), we found no significant pattern nor any evidence of systematic error.


\section{Prediction of change blindness}

\subsection{Observer variability and consistency}

\begin{figure}[!ht]
	\begin{center}
        \includegraphics[width=1.03\linewidth]{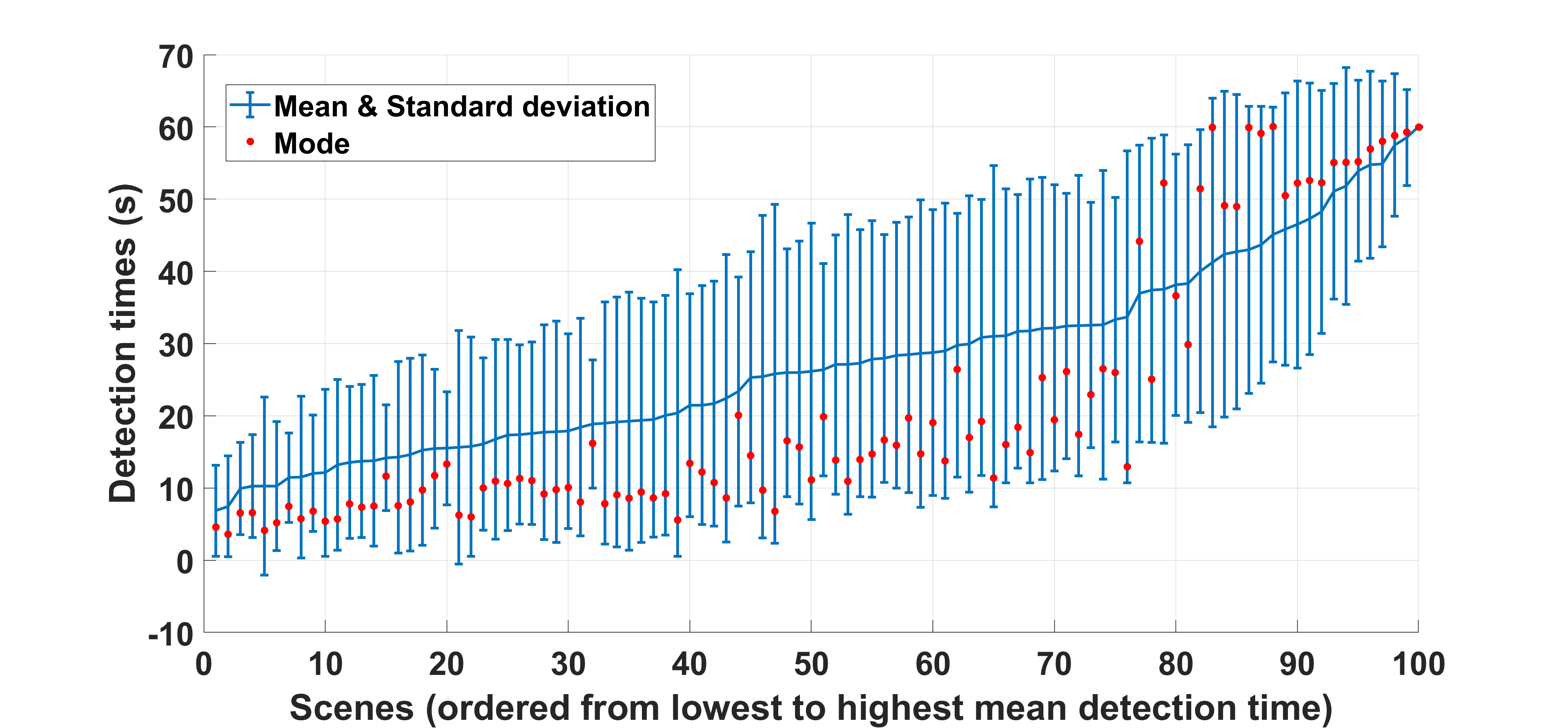} 
        \caption{Average (blue line), mode (red dots) and standard deviations (bars) of individual $T_{o,i}$. Notice that, when projecting all the red dots on the y-axis, there are two distinct clusters of scenes separated by a region of low density around 35s.}
    \label{fig:IOV}
	\end{center}
\end{figure}

\subsubsection{Intra-observer variability}

Change blindness is only temporary, which makes the associated intra-observer variability difficult to measure. If a person finds a change in a given image pair, any subsequent exposure to the same pair will almost certainly result in a faster detection. We can, however, analyse how consistently an observer performs with respect to the majority of others. If an observer is systematically slower or faster than all others, this indicates that the variability of this particular observer is smaller than inter-observer variability.

For a given $o$, we measured how systematically $T_{o,i}$ was either below or above the mode over all observers noted $\hat{T_{i}}$, (see \ref{sec:iov}). A sign test revealed that all of them were consistent in that regard ($p<0.01$), i.e. each observer is either systematically slower or faster than most. More specifically, 32\% of observers were systematically in the fastest 10\% while 8\% of them were systematically in the slowest 10\% (sign test, $p<0.01$). This suggests that, on our data, intra-observer variability is small compared to inter-observer variability.

\subsubsection{Inter-observer variability}
\label{sec:iov}

As depicted in Figure \ref{fig:IOV}, observers tend to agree most on which images are the easiest and most challenging ones. The standard deviations of DTs are the smallest at both ends of the graph.

By studying the consistency of observers to be either slower or faster than the majority of all other participants we found two emerging clusters. The cluster of 'slow' observers is of size 40, while the cluster of 'fast' observer is of size 20. It is particularly noteworthy that the average ages are respectively 38.6 and 29.1 years old. The significance of this difference was confirmed by an unequal variances $t$-test ($p<0.001$). This suggests a significant effect of age in the recorded detection times.

Furthermore, the distribution of DTs was found non-normal for all $\mathbf{p}_i$ (Kolmogorov-Smirnov test at the 5\% confidence level).
A univariate kernel density estimate (with bandwidth selection based on Silverman's rule of thumb) was used to estimate the probability density function of each $\mathbf{p}_i$. All estimates showed one dominant peak, at least twice as high as the next significant peak (if any). The distribution of these dominant modes over the 100 pairs is depicted in Figure \ref{fig:IOV}. As opposed to the mean value $\bar{T_{i}}$, $\hat{T_{i}}$ produces two clearly distinct clusters separated by the global density minimum at around 36 s. This led us to the conclusion that the mean DT is not the most appropriate statistic to represent the degree of change blindness engendered by a given pair (it was used for example in \cite{MA13}). Rather, we chose the dominant mode $\hat{T_{i}}$ as target value for our regression analysis (see results section). The two naturally emerging clusters $\mathbf{C1}$ and $\mathbf{C2}$ were then used as ground truth for our classification analysis:

\begin{equation*}
 \begin{aligned}
 & \mathbf{C1}=\left\{\mathbf{p}_i|i:\hat{T_{i}} < T_{\text{crit}}\right\} \\
 & \mathbf{C2}=\left\{\mathbf{p}_j|j:\hat{T_{j}} \geq T_{\text{crit}}\right\}
\end{aligned}
\end{equation*}

\noindent where $T_{\text{crit}}$ is the critical time (minimum density of DT modes): 36 s.

We also looked at the effect of experience on performance. A significant linear correlation ($r=0.41$, $p<0.001$) was measured between $\hat{T_{i}}$ and the average position of $\mathbf{p}_i$ within the sequence over all observers. This confirms that short-term experience partially predicts decision times.


\subsection{Proposed model}

In this paper, our aim is to create a very simple model that is intuitive, computationally efficient and relies on as few parameters as possible. The proposed model is based on three features:

\begin{itemize}
    \item \textbf{Change Magnitude} ($f_{\text{CM}}$),
    \item \textbf{Salience Imbalance} ($f_{\text{SI}}$),
    \item \textbf{User Experience} ($f_{\text{UE}}$).
\end{itemize}

Note that all stimuli are first converted to the LAB2000HL colour-space \cite{lissner2012toward} before feature extraction.




Change magnitude ($f_{\text{CM}}$) is calculated in terms of pixel colour difference (i.e. the Euclidean distance in LAB2000HL). Only the changed pixels are considered, which effectively means that the size of the change is not accounted for (as opposed to averaging over all pixels within the image). In fact, the information pertaining to the size of the changed region is partially carried by $f_{\text{SI}}$.

\begin{equation}
    f_{\text{CM}}(\mathbf{p}_i) = \frac{1}{\#D}\sum_{ D}\Delta E_{\text{00HL}}\left[\text{s}_i(x,y), \text{s}_i^\ast(x,y)\right]
\end{equation}

\noindent where

\begin{equation}
    D = \left\{\{x,y\}|\text{s}_i(x,y) \neq \text{s}_i^\ast(x,y)\right\}
\end{equation}

\noindent and $\text{s}_i(x,y)$ represents the pixel at spatial coordinates $x$ and $y$ in $\text{s}_i$.

Salience imbalance is calculated based on the approach proposed in \cite{HOU12}: the Hamming distance between the signs of Discrete Cosine Transform (DCT) coefficients of $\text{s}_i$, averaged over its three colour channels:

\begin{equation}
    f_{\text{SI}}(\mathbf{p}_i) = \frac{1}{3N}\sum_{k} d_{\mathbf{H}}\left[\mathbf{sign} \left(\text{DCT}_k(\text{s}_{i})\right),\mathbf{sign} \left(\text{DCT}_k(\text{s}_{i}^\ast)\right)\right]
    \label{eq1}
\end{equation}

\noindent where $d_{\mathbf{H}}\left(A,B\right)$ is the Hamming distance between vectors A and B, $N$ is the number of pixels in $\text{s}_i$ and $\text{DCT}_k(\text{s}_{i})$ denotes the discrete cosine transform of the $k$-th channel (either L$_{\text{00HL}}$, a$_{\text{00HL}}$ or b$_{\text{00HL}}$) of $\text{s}_i$ (same for $\text{s}_i^\ast$). Note that $\text{DCT}_k(\text{s}_{i})$ is in vector form in Eqn. (\ref{eq1}).

Individual user experience is represented by the index of the image pair within the random sequence generated for each observer (in other words: the number of previously seen examples plus one). The collective experience $f_{\text{UE}}$ associated with the mode of DTs is calculated as the squared average over all observers.

\begin{equation}
    f_{\text{UE}} = \left(\frac{1}{N_o}\sum_{o=1}^{N_o} \textit{id}_{i,o} \right)^2 
\end{equation}

\noindent where $\textit{id}_{i,o}$ is the index of image pair $\mathbf{p}_i$ in the sequence generated for observer $o$ and $N_o$ is the total number of valid observers (60).

The predicted detection time is eventually obtained via multivariate linear regression:

\begin{equation}
    \tilde{T_{i}} = b_1f_{\text{CM}}+b_2f_{\text{SI}}+b_3f_{\text{UE}}
\end{equation}

\noindent where $b_k$ ($k=1\dots 3$) are the only three parameters of the proposed model. In our experiments, these parameters were trained on a portion of the dataset (see next section). Note that we tried different pooling strategies and found that linear regression gives the best results overall on our data. We also considered a variety of other low-level features such as visual complexity \cite{rosenholtz2007measuring,mack2004computational}, alternative salience detection models or image-difference features \cite{le2017can}. We found that they did not help improve the model's performance.


\section{Results}

We analyse the performance of the proposed model in terms of regression and classification of detection times.

\subsection{Regression}

\begin{figure}[!ht]
	\begin{center}
            \begin{tabular}{cc}
             \includegraphics[width=0.4\linewidth]{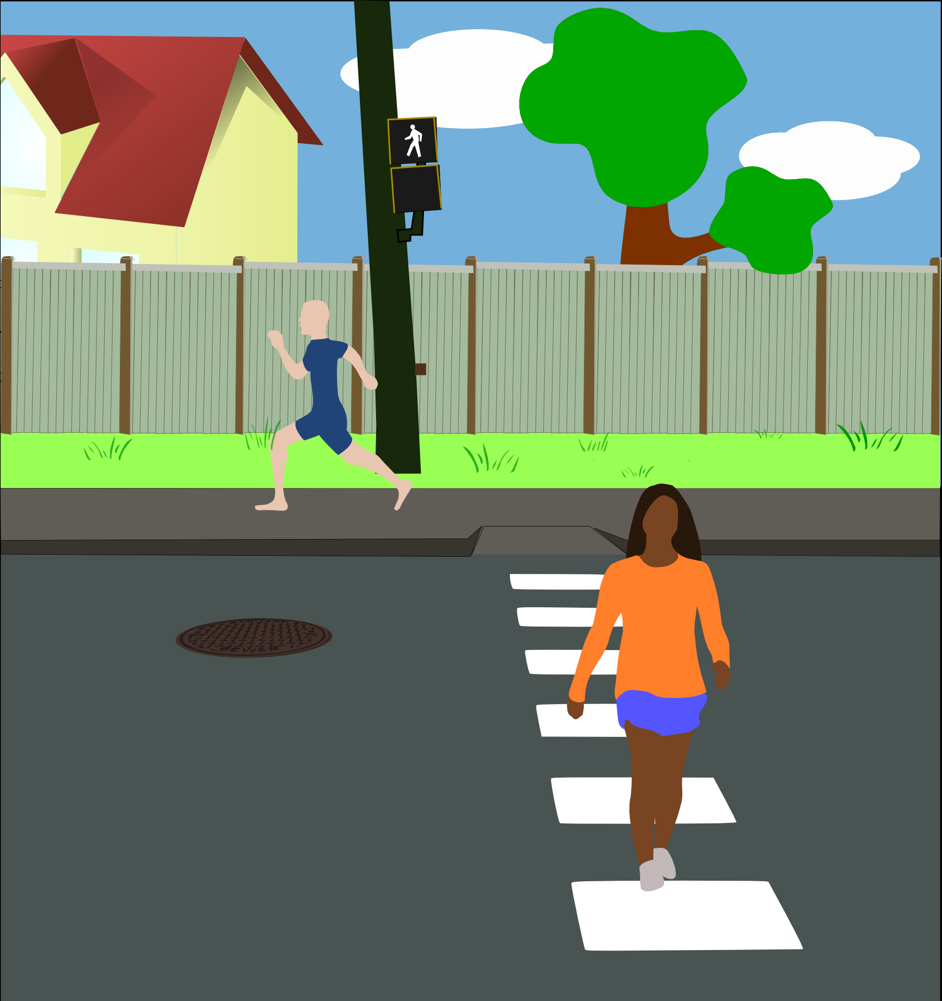} & 
             \includegraphics[width=0.4\linewidth]{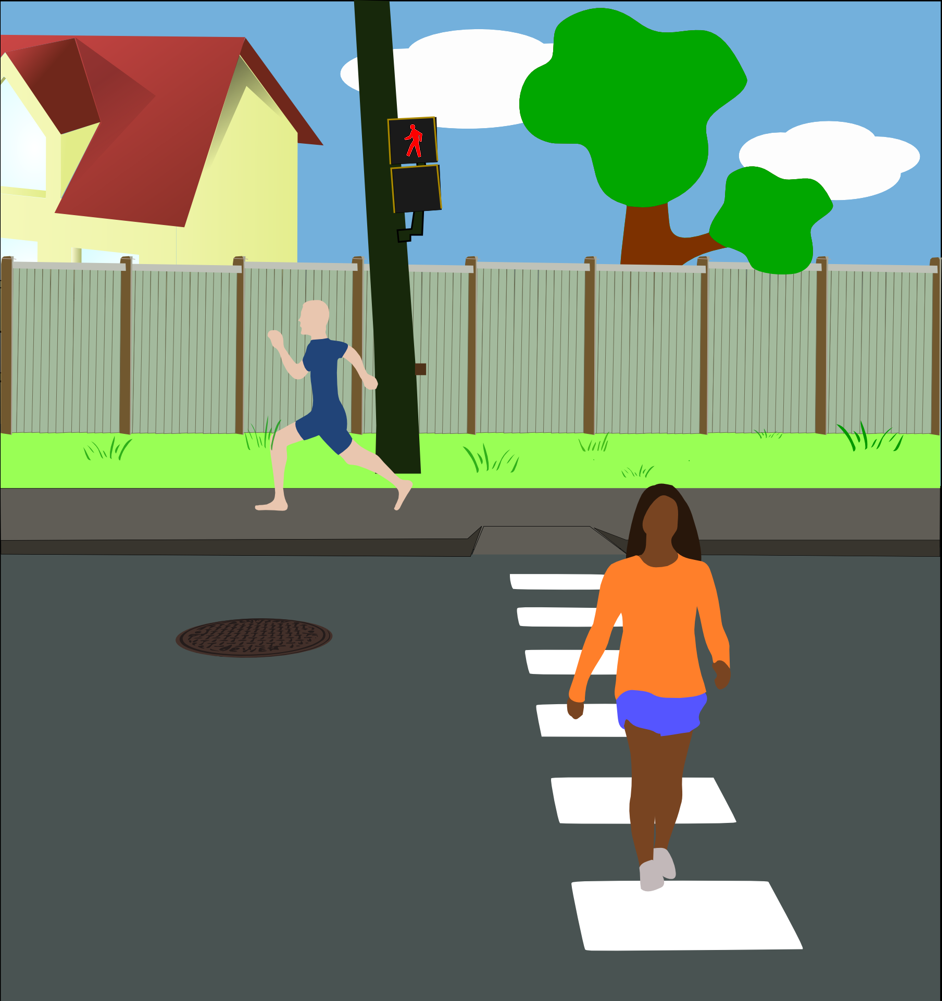} \\
             \includegraphics[width=0.45\linewidth]{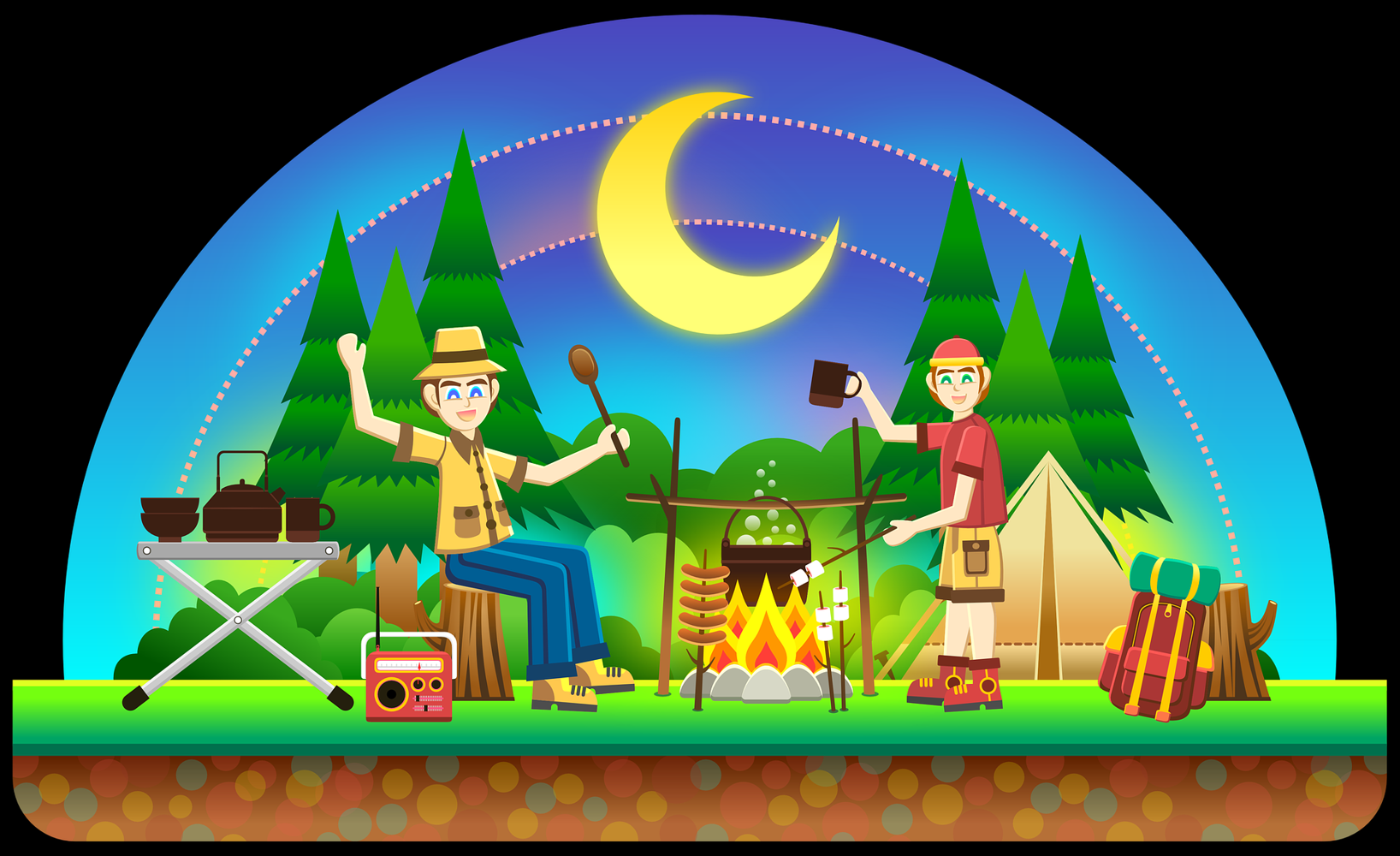} & 
             \includegraphics[width=0.45\linewidth]{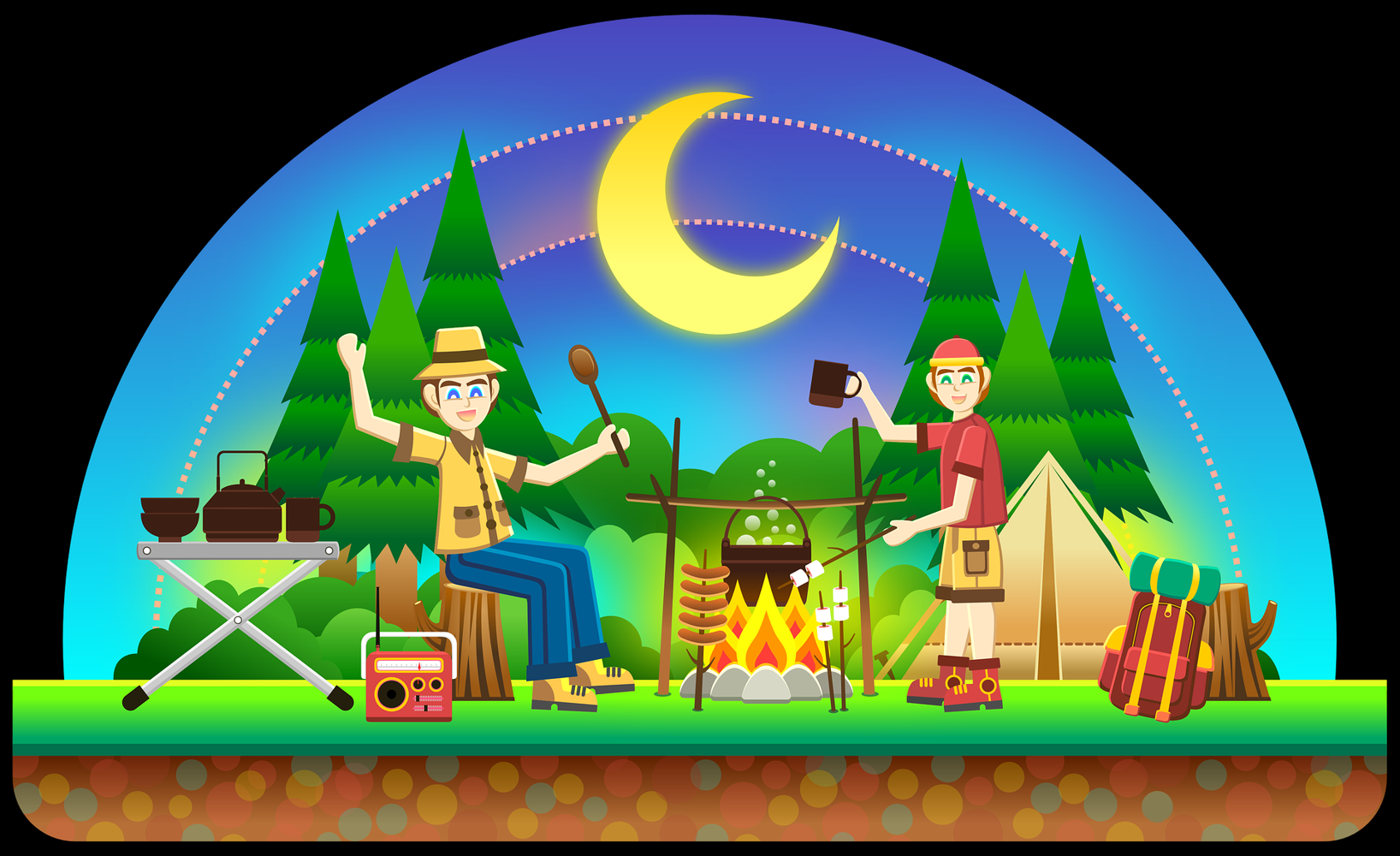} \\
            \end{tabular}
        \caption{Image pairs corresponding to the lowest (top) and largest (bottom) average DTs. The top scene is arguably subject to a top-down bias that draws attention to the traffic light, resulting in a faster detection. On the other hand, the bottom scene is particularly cluttered and the small magnitude of the change results in virtually no salience imbalance.}
    \label{fig:examples2}
	\end{center}
\end{figure}

We first give, in Table \ref{tab:res1}, the performance of individual features $f_{\text{CM}}$, $f_{\text{SI}}$ and $f_{\text{UE}}$ as well as age and visual complexity features \cite{rosenholtz2007measuring,mack2004computational}. The latter were calculated in two fashions: globally (accounting for all pixels) and locally (accounting only for pixels within a 20 pixel distance to the changed object/region, including the changed pixels themselves).

We then compare, in Table \ref{tab:res2}, linear regression to support vector regression, feedforward neural network and decision tree, as well as the model by Ma \etal{} \cite{MA13}, which we trained and tested on our data. Specifically, we trained the three parameters of Eqn. (2) in \cite{MA13} and all other parameters of the model were set as described in the paper. The support vector regression was trained with sequential minimal optimization and it was given standardised data as input for optimal performance. The neural network was trained with the Levenberg-Marquardt method and with Bayesian regularization. We gave it two neurons, which resulted in an average of 4.8 effective parameters. The decision tree was parametrised to have no more than 5 splits.

Cross-validation was performed by averaging of 100 unique random 70\%/30\% splits. Average Pearson and Spearman correlation coefficients were calculated by first converting the individual $r$ to Fisher's $z$ values, averaging them and then converting back to $r$. The significance of the difference between $r$ values was performed for each random split individually and averaged.

\begin{table*}[!ht]
	\begin{center}
		\caption{Prediction performance of individual features.}
		\begin{tabular}{|c|c|c|c|c|}
		    \hline
		    & \multicolumn{2}{c|}{Indiv ($T_{i}$)} & \multicolumn{2}{c|}{Mode ($\hat{T_{i}}$)}\\
		    \hline
		    & PLCC & SROCC & PLCC & SROCC \\
		    \hline
		    $f_{\text{CM}}$ & 0.28 & 0.30 & \textbf{0.37} & \textbf{0.45} \\
		    \hline
		    $f_{\text{SI}}$ (based on \cite{HOU12}) & -0.29 & -0.29 & \textbf{-0.39} & \textbf{-0.46} \\
		    \hline
		    $f_{\text{UE}}$ & 0.08 & 0.18 & \textbf{0.42} & \textbf{0.30} \\
		    \hline
		    age & 0.19 & 0.18 & / & / \\
		    \hline
		    Subband entropy (global) & 0.12 & 0.13 & 0.15 & 0.19 \\
		    \hline
		    Subband entropy (local) & 0.16 & 0.20 & 0.21 & 0.21 \\
		    \hline
		    Edge density (global) & 0.08 & 0.08 & 0.13 & 0.12 \\
		    \hline
		    Edge density (local) & 0.10 & 0.12 & 0.14 & 0.15 \\
		    \hline
		\end{tabular}
		\label{tab:res1}
	\end{center}
\end{table*}

\begin{table*}[!ht]
	\begin{center}
		\caption{Prediction performance of model (average of 100 random 70/30 splits). Correlation coefficients in bold font are not significantly different from each other column-wise ($p<0.01$)}
		\begin{tabular}{|c|c|c|c|c|c|c|c|}
		    \hline
		    & \multicolumn{3}{c|}{Indiv ($T_{i}$)} & \multicolumn{3}{c|}{Mode ($\hat{T_{i}}$)}\\
		    \hline
		    & PLCC & SROCC & RMSE & PLCC & SROCC & RMSE \\
		    \hline
		    Linear regression & \textbf{0.29} & \textbf{0.29} & 20.0 & \textbf{0.62} & \textbf{0.63} & 14.8 \\
		    \hline
		    SVR & \textbf{0.34} & \textbf{0.36} & 20.2 & \textbf{0.61} & \textbf{0.62} & 16.3 \\
		    \hline
		    NN & \textbf{0.36} & \textbf{0.35} & 19.3 & \textbf{0.55} & \textbf{0.57} & \textbf{15.6} \\
		    \hline
		    Tree & \textbf{0.40} & \textbf{0.37} & 19.0 & 0.45 & 0.41 & 17.7 \\
		    \hline
		    Ma \etal{} \cite{MA13} & 0.11 & 0.09 & 35.2 & 0.39 & 0.39 & 22.5 \\
		    \hline
		\end{tabular}
		\label{tab:res2}
	\end{center}
\end{table*}

Overall, the proposed model performs significantly better than any individual feature (including the model by Hou \etal{} \cite{HOU12}) and than the model by Ma \etal{} in predicting individual and mode detection times. Interestingly, features that capture visual clutter correlate poorly with detection times. These results also indicate that pooling the proposed features via linear regression gives results that are not significantly worse than via support vector machines, neural network or decision tree with an equal or greater number of parameters. Looking at the optimised parameters of the regression, we found that user experience is the most predictive parameter, followed by salience imbalance and finally change magnitude.

In terms of RMSE, the performance is fairly low all across the benchmark, with a minimum average of 14.8 s to predict mode detection times. This means that a more elaborated model is needed to predict decision times more accurately. Such a model should incorporate individual factors (e.g. cognitive abilities, contrast sensitivity) and high-level attributes (e.g. object importance and semantic categories) to account for top-down biases in visual search strategies. However, in terms of correlation, the proposed model performs well and can ranks the 100 image pairs in order of increasing difficulty with reasonable accuracy (SROCC 0.63), especially given its simplicity and the fact that it relies only on three parameters.

\begin{figure}[!ht]
	\begin{center}
            \begin{tabular}{cc}
             \includegraphics[width=0.45\linewidth]{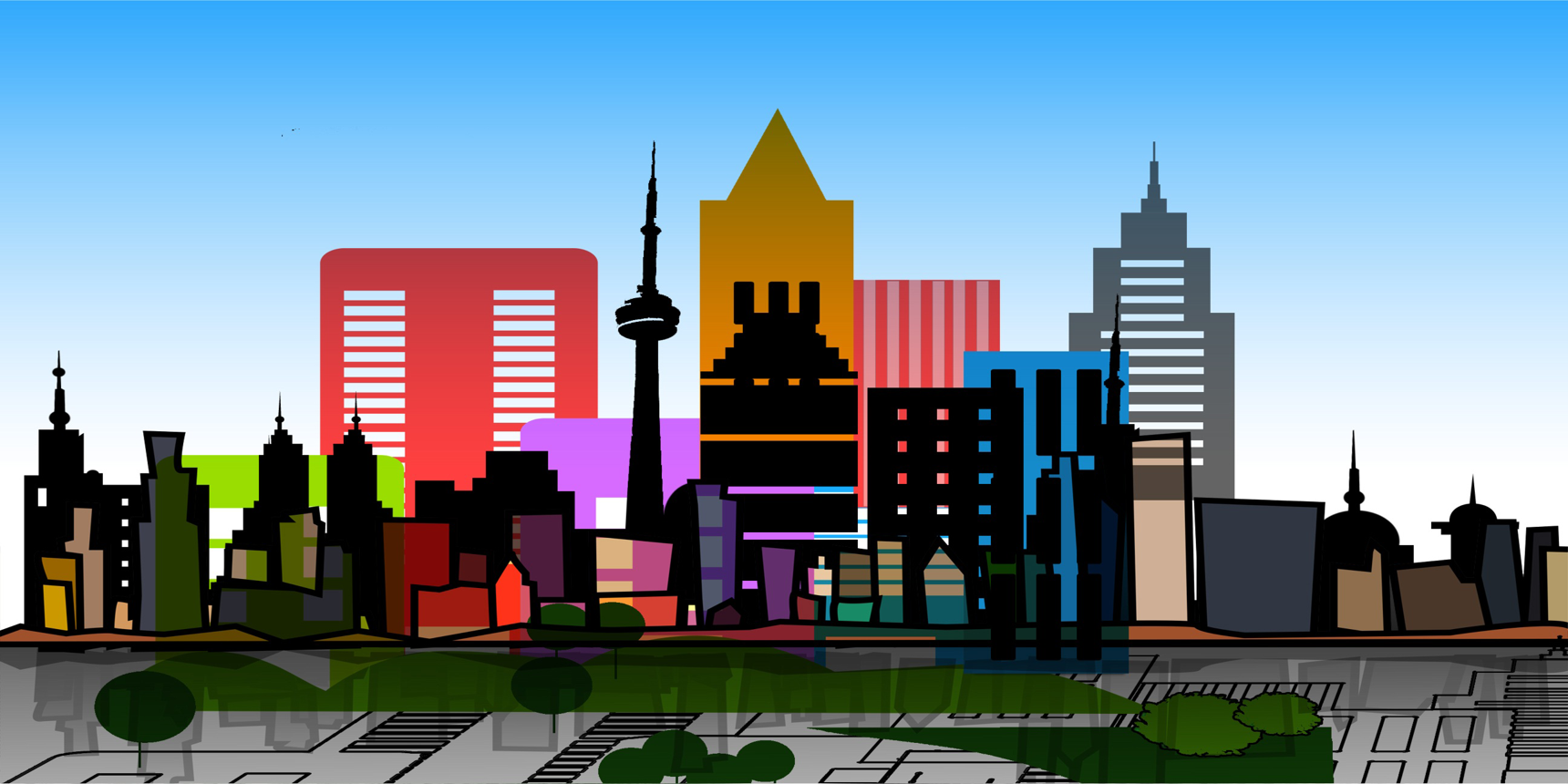} & 
             \includegraphics[width=0.45\linewidth]{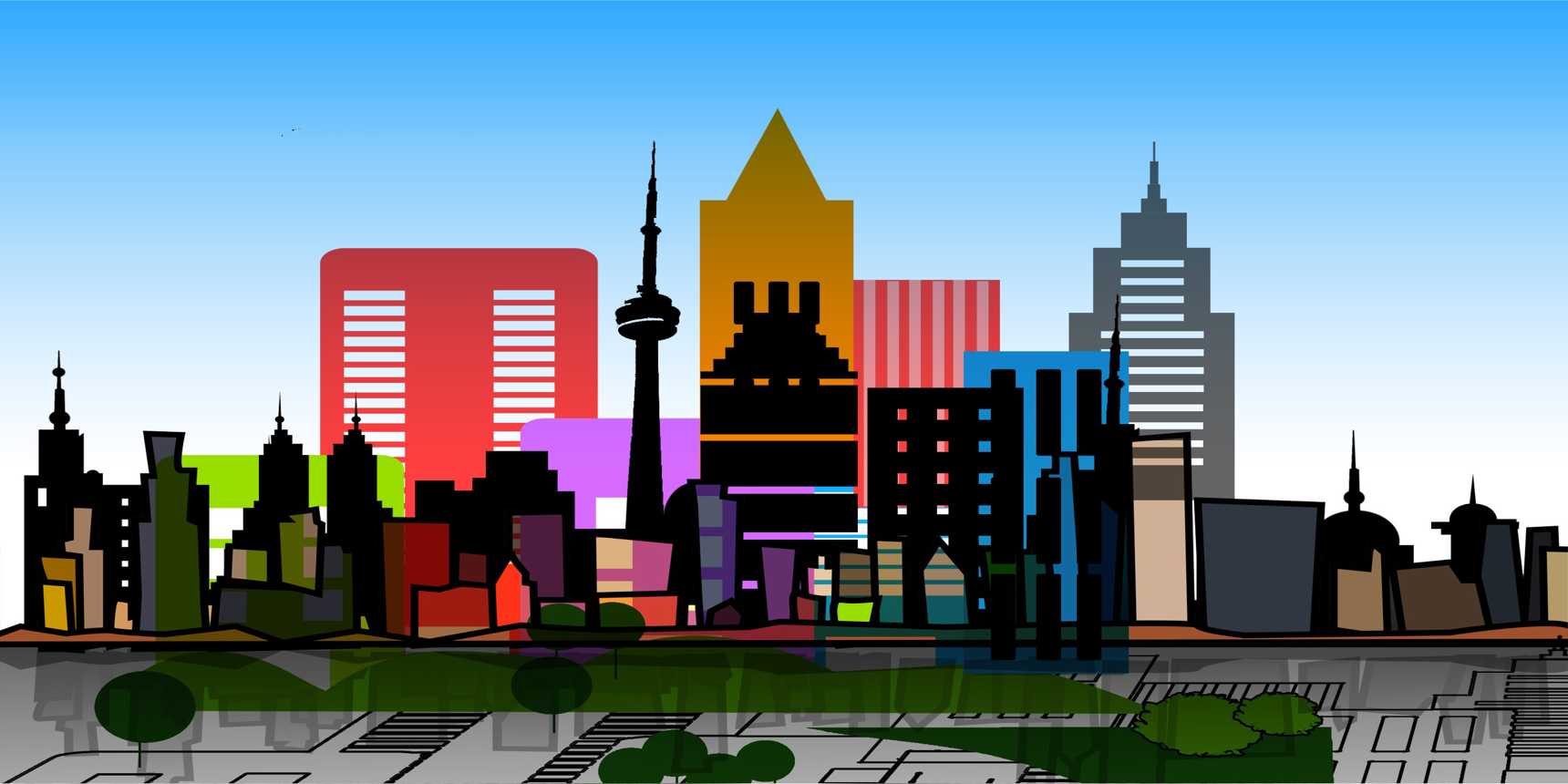} \\
             \multicolumn{2}{c}{$\hat{T_{i}} = 25.3 s$ $ f_{\text{CM}} = 0.20$ $ f_{\text{SI}} = 0.023$}\\
             \includegraphics[width=0.45\linewidth]{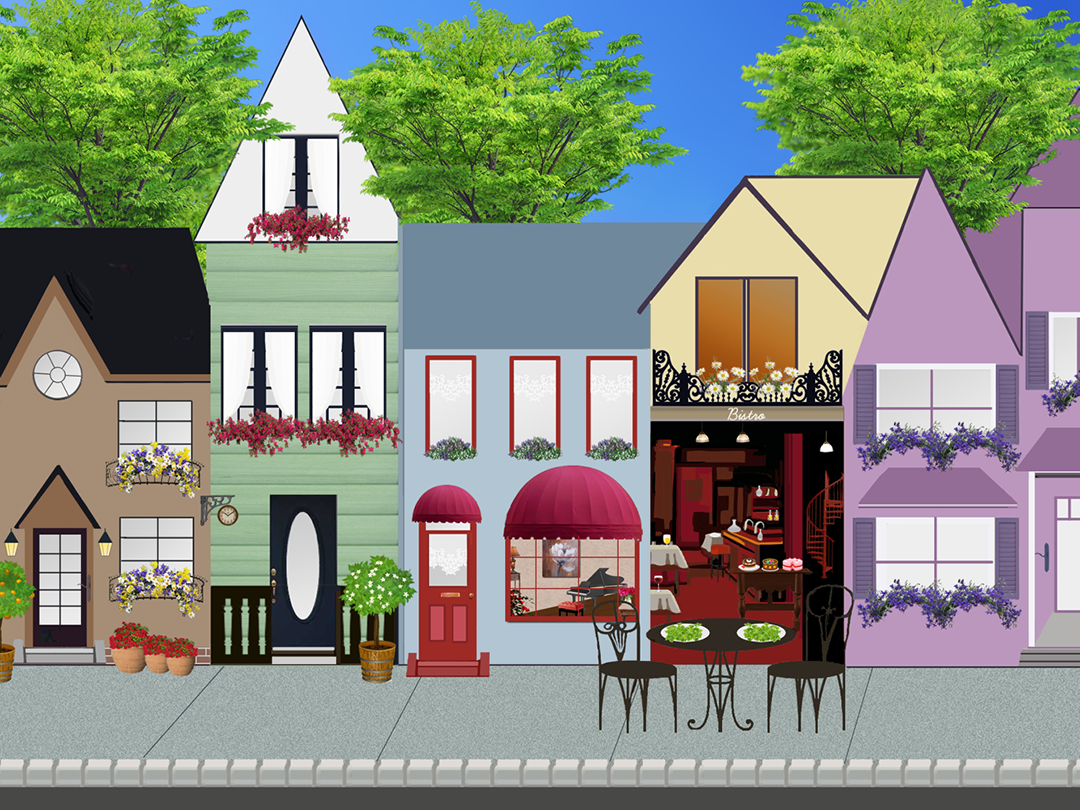} & 
             \includegraphics[width=0.45\linewidth]{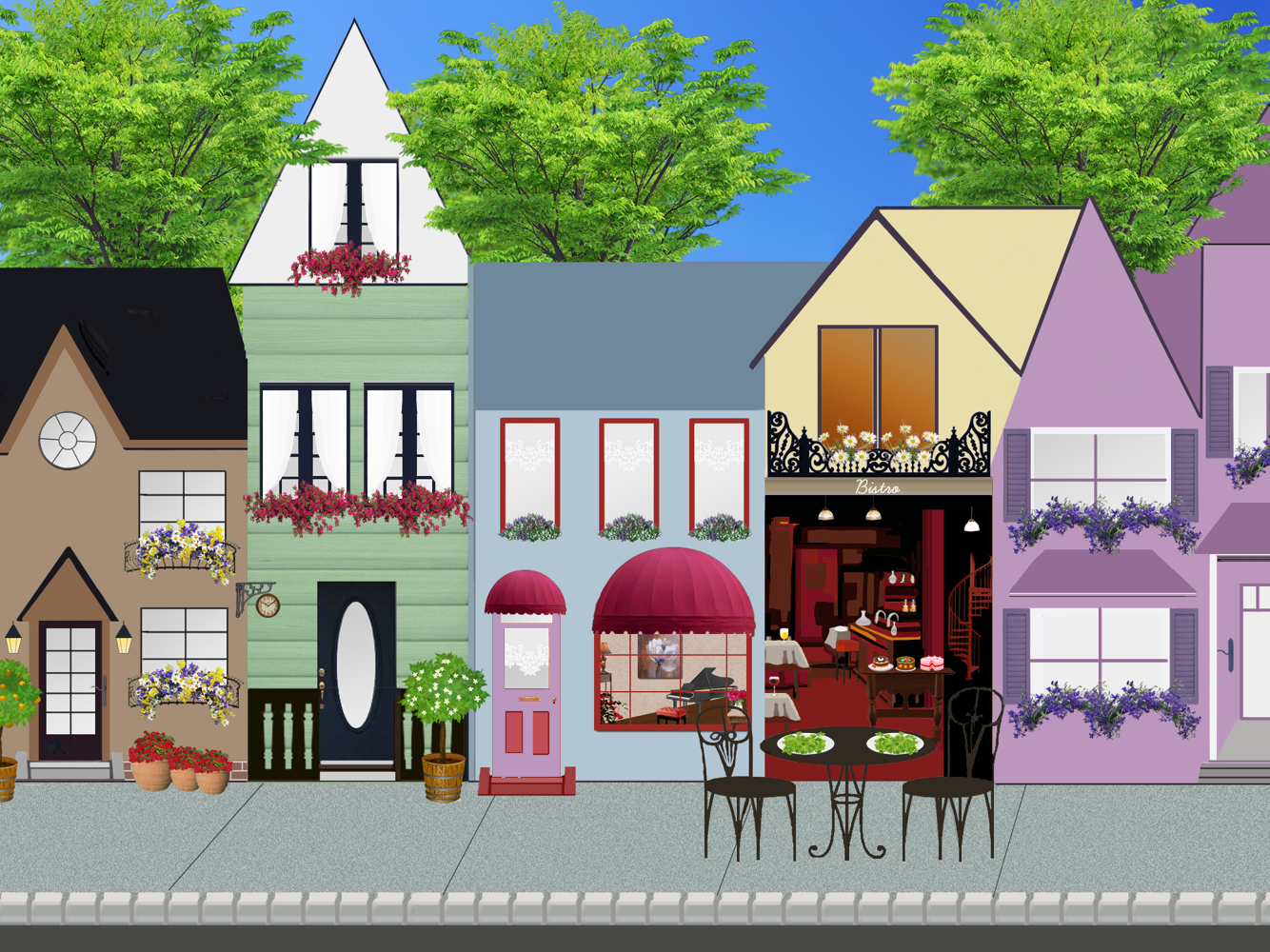} \\
             \multicolumn{2}{c}{$\hat{T_{i}} = 55.1 s$ $ f_{\text{CM}} = 0.38$ $ f_{\text{SI}} = 0.032$}\\
            \end{tabular}
        \caption{Image pairs corresponding to the best (top) and worst (bottom) prediction of mode detection times (in terms of RMSE). Solution, top: a rectangle just below the red building with vertical stripes, to the left: bottom: the central building's door.}
    \label{fig:best_worst}
	\end{center}s
\end{figure}

Figure \ref{fig:best_worst} shows image pairs for which the model gave the best and worst prediction of mode detection times (in terms of RMSE). Salience imbalance is nearly identical, but the bottom pair has a larger magnitude of change. However, the latter induces a longer detection time, likely due to more visual clutter. However, neither of the two clutter indices mentioned earlier were able to improve the performance in this case. We believe that a top-down approach to measuring visual clutter is needed to improve performance in such cases.


\subsection{Classification}

Image pairs were classified depending on whether $\hat{T(i)}$ was below or above 35 seconds ($\mathbf{C1}$ or $\mathbf{C2}$). We used Quadratic Discriminant Analysis to identify these two classes in the three-dimensional space spanned by the model's features. A ten-fold cross-validation gave an average overall accuracy of 85\% to classify mode detection times. Similar performance was obtained with support vector machines and neural network-based classification. Binary tree classification was also implemented, primarily in order to gain insight into the classification process. The resulting tree, where the number of nodes was set to not exceed 4 (for interpretability) yielded an average 0.78 overall accuracy. It shows that the most difficult image pairs are those with either:

\begin{itemize}
\item a small $f_{\text{UE}}$ and a large $ f_{\text{CM}}$.
\item a large $f_{\text{UE}}$, large $f_{\text{CM}}$ and small $f_{\text{SI}}$.
\end{itemize}

Based on the decision tree, user experience ($f_{\text{UE}}$) is the most predictive feature, as previously noted.

\section{Conclusions}

Change blindness is a complex phenomenon that involves advanced cognitive functions responsible for instance for attention and visual short-term memory. Despite our efforts to reduce subjective biases to a minimum, there are still uncertainties as to the exact origins of the phenomenon. We introduced a very simple model based on user experience, change magnitude (colour difference) and salience imbalance. The model relies on three parameters only that can predict mode decision times with significantly better accuracy than individual features and than existing models. The model can be used to rank image pairs in terms of difficulty and to classify as either easy or difficult with an overall accuracy of 85\%.

A more advanced model of change blindness will need to predict visual search strategies and eye movements, as well as the probability of internal representation. However, this would require more parameter tuning and substantially more data for calibration/training. Arguably, there would also be limits to the accuracy to be obtained (in terms of either RMSE or correlation), due to the stochastic nature of eye movements and visual search strategies. Recent research on saccade and fixation prediction \cite{le2015saccadic} have shown that predicting individual patterns is very difficult, suggesting that inter-observer variability is not yet well understood. In our analysis, we found that the distribution of detection times does not follow a normal distribution and that it often has several modes and always a dominant one. We proposed to use the latter to represent detection times, rather than the mean, as commonly done in the literature.

Our results also confirmed that predicting change blindness is difficult. It is, however, of great scientific value for applications in visual quality assessment, signal compression and computer graphics.

\section{Acknowledgments}

We would like to thank Li-Qian Ma for providing the implementation of their model.

\bibliographystyle{IEEEbib}
\bibliography{references}

\end{document}